\pdfoutput=1

\documentclass[11pt]{article}

\usepackage[]{EACL2023}

\usepackage{times}
\usepackage{latexsym}

\usepackage[T1]{fontenc}
\usepackage{float}
\usepackage{booktabs}

\usepackage[utf8]{inputenc}

\usepackage{microtype}

\usepackage{inconsolata}

\usepackage{amsmath,amssymb,amsthm} 
\usepackage{graphicx}
\usepackage{subfigure}
\usepackage{atbeginend}

\newtheorem{dfn}{Definition}
\newtheorem{theorem}{Theorem}
\newtheorem{prf}{Proof}

\title{Theoretical Conditions and Empirical Failure of Bracket Counting on Long Sequences with Linear Recurrent Networks}

\author{Nadine El-Naggar 
  \\\And
  Pranava Madhyastha \\
  City, University of London \\
  United Kingdom \\
  \texttt{\{nadine.el-naggar,pranava.madhyastha,t.e.weyde\}@city.ac.uk} \\
  \And
  Tillman Weyde 
  \\}

\begin{document}
\AfterBegin{itemize}{\addtolength{\itemsep}{-.8mm}}
\maketitle
\begin{abstract}
Previous work has established that RNNs 
with an unbounded activation function 
have the capacity to count exactly. However, it has also been shown that RNNs
are challenging to train effectively and generally do not  
learn exact counting behaviour. 
In this paper, we 
focus on
this  problem by studying the simplest possible RNN, a linear single-cell network.  
We conduct a theoretical analysis of linear RNNs and identify conditions for the models to exhibit \emph{exact} counting behaviour. 
We provide a formal proof that these conditions are necessary and sufficient.
We also conduct an empirical analysis using tasks involving a Dyck-1-like Balanced Bracket language under two different settings. 
We observe that linear RNNs generally do not meet the necessary and sufficient conditions for counting behaviour when trained with the standard approach.
We investigate how varying the length of training sequences and utilising different target classes impacts model behaviour during training and the ability of linear RNN models to 
effectively
approximate the indicator conditions.
\end{abstract}

\section{Introduction}

Recurrent Neural Networks (RNNs) have a long history of being used to solve a wide range of tasks involving sequential data. They were 
the most 
common choice for natural language processing, but have since been largely replaced by transformers in recent years. 
However, there has been a recent resurgence of interest in the theoretical aspects of RNNs, as seen in studies such as \citet{merrill2020formal}.
Another study found that RNNs with squashing and non-squashing (i.e. unbounded) activation functions exhibit qualitative differences in their counting abilities \cite{weiss-et-al-2018-practical}. 
This, along with 
the findings of \citet{el-naggar-et-al-2022experiments}, 
suggests that even RNNs with unbounded activation functions struggle to learn accurate counting on very long sequences. 
It is therefore crucial to understand why RNNs, despite having the capacity, often fail to accurately count in practice.

In this study, we examine the behaviour of the simplest form of RNNs: a linear single-cell RNN. 
%
Our goals are to:
a) theoretically identify the necessary conditions for a linear RNN to have the ability to count, and
b) explore how these conditions relate to the empirical behaviour of trained linear RNN models.
%
The primary contributions of this paper are:
    a) we identify two conditions that indicate counting behaviour in linear RNNs;
    b) we prove that these indicator conditions are necessary and sufficient for exact counting behaviour to be achieved in linear RNNs;
    c) we then show empirically that linear RNNs generally do not learn exact counting and do not meet the indicator conditions; and finally, 
    d) we show empirical relationships between the length of the training sequences 
    and the indicator value distributions. 

\section{Related Work}

The success of deep learning sparked a renewed interest in research into the understanding of the theoretical properties of neural networks. 
It has been known for long that RNNs are Turing-complete \cite{siegelmann-sonntag-1992-computational}. 
However, \citet{weiss-et-al-2018-practical} showed that there are different classes of RNN architectures with respect to counting capacity when using finite precision states. 
The relationship between RNNs and automata and formal languages has been investigated by \citet{merrill2019sequential} and a
formal hierarchy of counter machines has been developed by \citet{merrill2020formal}. 
This analysis is often based on ``saturating'' the network, i.e. replacing sigmoids with step functions, so that neural activations become binary, allowing for simpler analysis.  

In practice however, activations in neural networks use a wide variety of values and systematic behaviour like counting or even parsing is often not observed, which has been discussed for over 30 years since \citet{fodor1988connectionism}. 
More recently, \citet{lake2017generalization} identified a specific lack of systematic behaviour and devised SCAN, a synthetic language processing task that standard RNNs fail on. 
A traditional approach for processing the hierarchical structure of language is to use a stack and a number of neural stack versions have been introduced, such as those by \citet{joulin2015inferring}, \citet{grefenstette2015learning} and \citet{chen2020compositional}, which performed well on SCAN.  
\citet{suzgun2019memory} use stack-augmented RNNs to learn Dyck-2 languages. 
\citet{mali2021investigating} developed methods to achieve more stable behaviour of neural stacks, achieving good, but not perfect, performance on longer sequences up to 160 tokens. 

The generalisation of formal language tasks to very long sequences is not often addressed, as it requires an exact or near exact behaviour of the neural network in order avoid accumulation of errors, e.g. when counting. 
\citet{suzgun2019lstm} claims that LSTMs can learn to count, but did not test for sequences of length greater than 100. 
\citet{weiss-et-al-2018-practical} reported that ReLU RNNs, which were generally hard to train, and even LSTMs which are easier to train on counting tasks, did fail for sequences of several hundred tokens. 
Similarly, in our previous work \citep{el-naggar2022exploring} we show that almost all ReLU RNNs and LSTMs fail on sequences of length 1000.

These studies generally indicate that RNNs do not reach a configuration that enables exact counting. 
It is not clear what the general conditions are for an RNN to perform exact counting, which is necessary for developing a deeper understanding of the behaviour of RNNs. 
We start to address this question here by studying the case of a linear single-cell RNN.

\section{Counting Behaviour in Linear RNNs}
In this section we formally define the Balanced Bracket Language, Balanced Bracket Counter and Linear Recurrent Network and we identify conditions for the network weights that indicate that the Linear Recurrent Network will behave as a Balanced Bracket Counter. 
We base our counter definitions on the General Counter Machine (GCM) as defined by \citet{DBLP:journals/corr/abs-2004-06866}, which we also listed for convenience in Appendix~\ref{apx:definition}. 




The GCM is defined by a vocabulary $\Sigma$, finite set of states $Q$, initial state $q_{0}$, counter update function $u$, state update function $\delta$, and acceptance mask $F$. 
Some components, such as states, can be empty. 
The counter computation also uses a zero check function.
An input string $x$ is processed by the counter one token $x_{t}$ at a time. 
The counter update function $u$ is used to update the counter value \textbf{c} with integer increments $(+m)$.
The counter updates are dependent on the current input token, the current state, and a finite mask of the current state.
In our setting, a counter machine is said to accept a sequence if $\textbf{c} = 0$ after the whole sequence is has been processed. 
A counter machine $M$ is said to accept a language $L$ if $M \text{accepts}\, s \iff s \in L$. 
 


We focus on sequences consisting of one type of bracket:
\(\Sigma = \)\{`(', `)'\}.

\begin{dfn}
\normalfont
(Balanced Bracket Language $BB$) \\
The Balanced Bracket Language $BB$ is defined as
$$BB = \{s \in \Sigma^* | \text{ count( `('}, s) = \text{count( `)'}, s) \}.$$
\end{dfn}
The order of the opening and closing brackets does not matter for the $BB$ language, only that the number of opening and closing brackets in a sequence is equal overall.
Dyck-1 sequences are a special case of $BB$ sequences where the number of closing brackets is never greater than the number of opening brackets at any point in the sequence, i.e. for all prefixes.

Our focus is on the counting abilities of single-cell linear RNNs. 
These networks do not have the capacity to accept Dyck-1 sequences from the universe of all possible sequences, because they would need to treat negative counts differently from positive activations to distinguish correctly ordered from incorrectly ordered bracket sequences. 
However, that is not possible with a single linear neuron, and 
additional mechanisms would be needed to fully accept a Dyck-1 language from the entire universe of possible sequences.

Previous work, such as \citet{suzgun2019lstm}, who train their single-cell RNN models to learn Dyck-1 languages only use valid Dyck-1 sequences in their datasets, where there are never any excess closing brackets at any point in the sequences.
This seems unnecessarily limiting, however. 
Therefore, we use the $BB$ language which can be fully accepted using a single-cell linear RNN.

\begin{dfn}
\normalfont
\label{dfn:BalancedBracketCounter}
(Balanced Bracket Counter) \\
A General Counter Machine is a Balanced Bracket Counter iff it accepts $BB$.
\end{dfn}

\begin{dfn}
\normalfont
\label{dfn:linear_recurrent_machine}
(Linear Recurrent Network) \\
A Linear Recurrent Network (LRN) is a network which receives an input $x_t$ at every timestep $t$, which is used along with the activation 
from the previous timestep $h_{t-1}$ and weights $W$, $U$, and $W_{b}$ to produce activation $h_{t}$, which is then passed on to the next timestep with the
update function: 
$$h_{t} = Wx_{t}+Uh_{t-1}+W_{b} .$$
\end{dfn}

Here, $x_{t}$ is a one-hot-encoded input token, 
an LRN is similar to a stateless counter machine if  we apply a zero check $z$ function to $h_t$ after processing the last input. 
A counter based on the LRN deviates from the definition by \citet{DBLP:journals/corr/abs-2004-06866} in that:
\begin{itemize}
    \item The counter value $\textbf{c}$ corresponds to $h_t$ (it is the only value that propagates from one time step to the next), which is real  instead of integer,
    \item The results of the update function ($+m$ in the Counter Machine) are real numbers, specifically:
    \begin{itemize}
    \item[]
    $a = W x_t + W_b$ if $x_{t} = $ `(' , and
    \item[]
    $b = W x_{t} + W_{b}$ if $x_{t} = $ `)' .
    \end{itemize}
\end{itemize}

We use a single-cell LRN for bracket counting. 
As a result, $W$ is a vector of the same dimensionality as $x_{t}$ and $U$ and $W_{b}$ are scalars, as well as  $m$, $\textbf{c}$, and $h_t$.

In Theorem~\ref{theorem:counting_indicators_linear}, we relate Balanced Bracket Counter behaviour of a LRN to specific conditions on its \emph{weights}.
We define two indicator conditions and show that they are necessary and sufficient for exact counting behaviour to be achieved in a LRN.

\begin{theorem}
\normalfont
\label{theorem:counting_indicators_linear}
(Linear RNN Counting Indicators) \\
The following two indicator conditions are necessary and sufficient for a Linear Recurrent Network to 
accept the Balanced Bracket Language $BB$.
\begin{enumerate}
    \item $\frac{a}{b} = -1$ \hspace{3mm} (AB ratio)
    \item $U=1$ \hspace{3mm} (recurrent weight).
\end{enumerate}
\end{theorem}

\begin{prf}
\normalfont
We prove that the counting indicator conditions in Theorem~\ref{theorem:counting_indicators_linear} 
are necessary and sufficient to 
accept the 
Balanced Bracket Language 
with a Linear Recurrent Network.
We first prove that the conditions are necessary (Part 1) and then that they are sufficient (Part 2).
\\
\begin{table}[!h]
    \centering
    \resizebox{0.95\linewidth}{!}{
    \begin{tabular}{cccc}
        \toprule
         \textbf{Case}&\textbf{Input} & \textbf{Output ($h$)} & \textbf{Findings}  \\ \midrule
         1&`('&$h_{1}\neq 0$ & $a \neq 0$ \\ 
         2 & `)' & $h_{1}\neq 0$ & $b \neq 0$\\ 
         3 & `()' & $h_{2} = 0$ & $b=-Ua$ and $U \neq 0$\\ 
         4& `((' & $h_{2} \neq 0$ & $U\neq -1$ \\ 
         5 & `(())' & $h_{4} = 0$ & $b+Ub+U^{2}a +U^{3}a = 0$\\ 
         6 & `()()' & $h_{4} = 0$ & $b+Ua + U^{2}b + U^{3}a=0$\\ \bottomrule
         
    \end{tabular}
    }
    \caption{Input sequences used to derive the indicator conditions from Theorem \ref{theorem:counting_indicators_linear}.}
    \label{tab:linear_cases}
\end{table} 
\textbf{Part 1:}
We prove that the counting indicator conditions in Theorem~\ref{theorem:counting_indicators_linear} 
are satisfied if a Linear Recurrent Network 
accepts the Balanced Bracket Language 
by using different input sequences (Table \ref{tab:linear_cases}), from which we derive the indicator conditions.
If a Linear Recurrent Network 
accepts the Balanced Bracket Language, 
then equal numbers of opening and closing brackets result in an output activation $h_{t}=0$, otherwise, $h_{t}\neq 0$.
This is equivalent to zero check function $z(h_{t})$ 
yielding 0 or 1.
Therefore, we will not include the zero-check function in the following derivation.

\begin{enumerate}
    \item[] \textbf{Case 1: } 
    $seq = $`('$, h_0 = 0, h_{1}\neq 0$\\
    $h_{1} = a + Uh_{0} = a$\\
    $\therefore a \neq 0$ 
        
    
    \item[] \textbf{Case 2:}
    $seq = $`)'$, h_0 = 0,  h_{1}\neq 0$\\
    $h_{1} = b + Uh_{0} = b$\\
    $\therefore b \neq 0$ 
    \item[] \textbf{Case 3:} $seq = $ `()', $h_{2} = 0$\\
    $h_{2} = b+Ua$\\
    $b+Ua = 0$\\
    From cases 1,2: $a\neq 0$ and $b\neq 0$\\
    $\therefore b = -Ua$, and $U\neq 0$ 
    \item[] \textbf{Case 4:} $seq = $`((', $h_{2} \neq 0$\\
    $h_{2} = a+Ua$\\
    $a + Ua \neq 0$\\
    $\therefore U \neq -1$ 
    \item[] \textbf{Case 5: }$seq = $`(())', $h_{4} = 0$\\
    $h_{3} = b + Uh_{2} = b + U(a + Ua)$\\
    $h_{4} = b + Uh_{3} = b + U(b + U(a + Ua)) \\\therefore h_{4}= b + Ub + U^{2}a + U^{3}a=0$ 
    \item[] \textbf{Case 6: }$seq = $`()()', $h_{4} = 0$\\
    $h_{3} = a + Uh_{2} = a + U(b + Ua)$\\
    $h_{4} = b + Uh_{3} = b + U(a + U(b + Ua)) \\\therefore h_{4}= b + Ua + U^{2}b + U^{3}a=0$ 

\hspace{5mm} Combine the findings from cases 5 and 6.\\
$b + Ub + U^{2}a + U^{3}a = b + Ua + U^{2}b + U^{3}a$


\hspace{5mm} Subtract $b + U^{3}a$ from both sides and divide both sides by $U$\\
$b+Ua = a + Ub$


\hspace{5mm} Rearrange and factorise\\
$b-a = U(b-a)$

As a result, we get 2 possible situations:
\begin{enumerate}
    \item $U = 1$, which implies $a=-b$ by case 3
    \item $a=b$, where by cases 1 and 2 we know $a \neq 0$ and $ b \neq 0$, and $U = -1$ follows from case 3, which  contradicts case 4

\end{enumerate}

$\therefore U = 1$ and $a=-b$, i.e., the counter indicator conditions listed in Theorem~\ref{theorem:counting_indicators_linear} hold, if a Linear Recurrent Network 
accepts the Balanced Bracket Language.\\

\end{enumerate}
\textbf{Part 2:}
We prove by induction that if the counting indicator conditions listed in Theorem \ref{theorem:counting_indicators_linear} hold, a Linear Recurrent Network 
accepts the Balanced Bracket Language.

Each sequence consists of $n$ opening brackets and $m$ closing brackets, and the input token $x_{k}$ is either `(' or `)'.\\
\textbf{Base Case: $k=1$}
\begin{itemize}
    \item $x_{1} = $`(', $n=1$ and $m=0$:\\
    $h_{1} = a+Uh_{0}$\\
    $h_{1} = a$
    \item $x_{1} = $`)', $n=0$ and $m=1$:\\
    $h_{1} = -a + Uh_{0}$\\
    $h_{1} = -a$
\end{itemize}
For $n$ opening and $m$ closing brackets, the following equation satisfies the base case, and is therefore our induction hypothesis:\\
$$h_{k} = (n-m)\times a$$ 
We assume that this is true for sequences of length $k$ consisting of $n$ opening brackets and $m$ closing brackets.
We prove by induction that if this holds for sequences of length $k$ tokens, it holds for sequences of length $k+1$ tokens.
In our induction step, we use once $x_{k+1} = $`(' and once $x_{k+1} = $`)'.
\\
\textbf{Induction Step:}
\begin{itemize}
    \item If $x_{k+1} = $`(':\\
    $h_{k+1} = ((n+1)-m)\times a$
    \item If $x_{k+1} = $`)':\\
    $h_{k+1} = (n-(m+1))\times a$
\end{itemize}
From the premise, we can derive that:\\
$
h_{k} = a+Uh_{k-1}$ if $x_{k} = $`(', and $h_{k} = -a+Uh_{k-1}$ if $x_{k} = $`)'
\begin{itemize}
    \item If $x_{k+1} = $`(':\\
    $h_{k+1} = a+h_{k}$\\
    Substitute $h_{k} = (n-m)\times a$\\
    $h_{k+1} = a+((n-m)\times a)$\\
    $\therefore h_{k+1} = ((n+1)-m)\times a$
    \item If $x_{k+1} = $`)':\\
    $h_{k+1} = -a+h_{k}$\\
    Substitute $h_{k} = (n-m)\times a$\\
    $h_{k+1} = -a+((n-m)\times a)$\\
    $\therefore h_{k+1} = (n-(m+1))\times a$
\end{itemize}
Therefore, we prove that if the counting indicator conditions listed in Theorem \ref{theorem:counting_indicators_linear} are satisfied in a Linear Recurrent Network, 
it 
accepts the Balanced Bracket Language.
\qed
\end{prf}

\begin{table*}[!htbp]
    \centering
      \resizebox{0.72\textwidth}{!}{
      \begin{tabular}{cccccl}
    \toprule 
    Classification & Train & Train 
    & 20 Tokens & 50 Tokens\\
    Experiment & Length & Avg(Min/Max) & Avg(Min/Max) & 
    Avg(Min/Max) \\
 \midrule
    Binary (without bias) & 2 &  100 (100/100)  
    & 69.2 (6.04/77.3) &  69.0 (66.7/72.7) \\
    Binary (without bias)  & 4  & 100 (100/100) 
    & 94.8 (94.7/95.3) & 89.3 (88.7/90.0) \\
    Binary (without bias) & 8 & 100 (100/100)  
    & 96.9 (94.0/100) & 92.7 (78.7/98.0) \\
    \midrule
    Ternary (without bias) & 2 & ~~90 (33.3/100) 
    & 55.6 (33.3/64.4) & 51.4 (33.3/60.0)\\
    Ternary (without bias) & 4 & 100 (100/100)
    & 79.5 (65.8/94.7) & 67.2 (66.7/68.0)\\
    Ternary (without bias) & 8 & 100 (100/100) 
    & 94.4 (67.1/100) & 85.7 (66.7/100)\\
    \midrule
    Binary (with bias) & 2& 100 (100/100) & 73.4 (63.3/100) & 72.4 (60.0/93.3)  \\
    Binary (with bias) & 4& 100 (100/100) & 95.3 (92.7/98.0) & 86.0 (77.3/90.7) \\
    Binary (with bias) & 8& 100 (100/100) & 95.2 (85.3/100) & 87.9 (70.0/98.0)  \\
    \midrule
    Ternary (with bias) & 2& 88.3 (66.7/100) & 58.0 (38.2/67.6) & 54.4 (43.6/67.5) \\
    Ternary (with bias) & 4& 97.9 (79.2/100) & 81.5 (64.4/100) & 68.0 (65.3/73.3) \\
    Ternary (with bias) & 8& 100 (100/100) & 95.9 (83.6/100) & 76.5 (65.3/100) & \\
    \bottomrule
  \end{tabular}
  }
    \caption{Accuracy metrics of our previous binary classification experiments without bias \cite{el-naggar-et-al-2022experiments}, ternary classification experiments without bias, and binary and ternary classification experiments with bias.}
    \label{tab:my_label}
\end{table*}

    


\section{Counting in Linear RNNs in Practice}
We conduct experiments to analyse the models and whether or not they satisfy the conditions defined in the previous section in training.
We use 2 classification tasks to evaluate our models.

\begin{figure*}[!h]
    \centering
    \subfigure[AB ratio binary (no  bias)]{\includegraphics[width=0.24\textwidth]{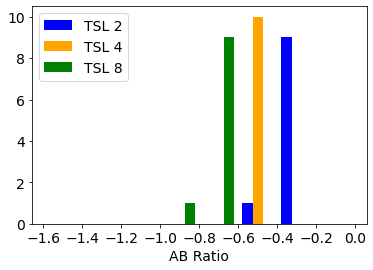}}
    \subfigure[U value binary (no  bias)]{\includegraphics[width=0.24\textwidth]{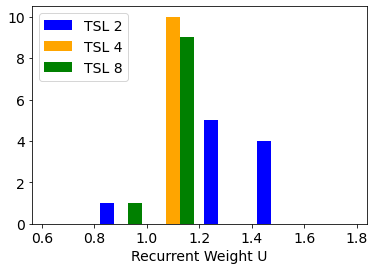}}
    \subfigure[AB ratio ternary (no  bias)]{\includegraphics[width=0.24\textwidth]{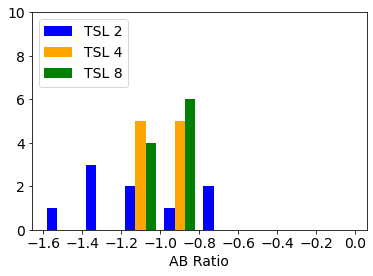}}
    \subfigure[U value ternary (no  bias)]{\includegraphics[width=0.24\textwidth]{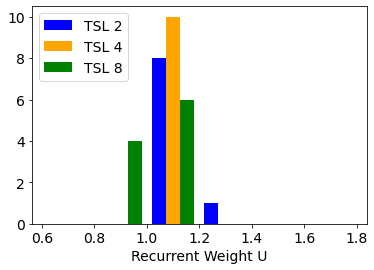}}
    \subfigure[AB ratio binary (bias)]{\includegraphics[width=0.24\textwidth]{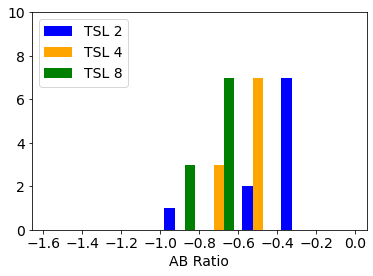}}
    \subfigure[U value binary (bias)]{\includegraphics[width=0.24\textwidth]{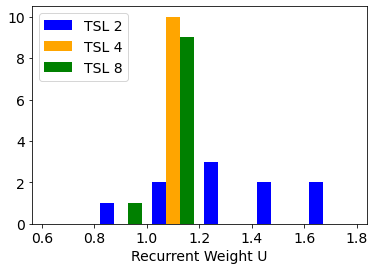}}
    \subfigure[AB ratio ternary (bias)]{\includegraphics[width=0.24\textwidth]{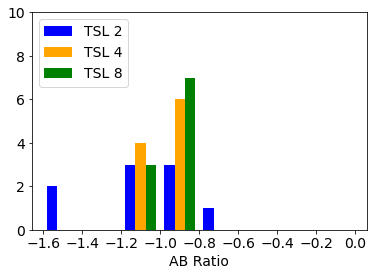}}
    \subfigure[U value ternary (bias)]{\includegraphics[width=0.24\textwidth]{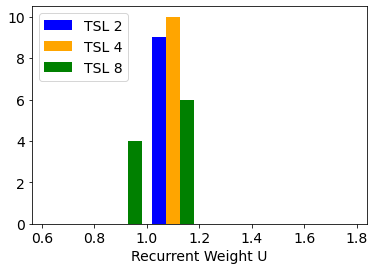}}
    \caption{Indicators after training on binary and ternary classification without biases (top) and with biases (bottom) with different Training Sequence Lengths (TSL).}
    \label{fig:histograms_with Bias}
\end{figure*}
\subsection{Task 1: Binary Classification}
We use the same task and model as in our previous work 
\citep{el-naggar-et-al-2022experiments}, i.e.,  a linear RNN without biases with a single output neuron with sigmoid  activation to classify the bracket difference of the sequence as $>0$ or $\leq 0$ (binary).
The absence of a trainable bias reduces the degrees of freedom in the model, and is equivalent to having a bias ($W_{b}$) value that is fixed to 0, hence simplifying the learning task. 
We also use the same models with trainable biases.
The models are trained with sequences of lengths 2, 4 and 8 tokens for 100 epochs in 10 runs.
The initial count value ($h_{0}$) has a value of 0 for every sequence.
We inspect the weights of our models and plot the distribution of the indicator values ($a/b$,$U$) of the trained models for each training set size in 
Figure~\ref{fig:histograms_with Bias}.
We observe that the models do not fulfill the indicator conditions, but they do approach the conditions as the length of the training sequences increases.
We observe that the distributions of the $a/b$ indicator have mean values above $-1$ but less so for longer training sequences.

\subsection{Task 2: Ternary Classification}
We also apply a ternary classification: $>0$, $=0$ or $< 0$.
We use the same model as in Task~1, except that instead of a single output neuron with a sigmoid output activation, we use 3 output neurons and a softmax output layer with bias, which is the minimal configuration that can achieve this task. 
We also use the same models with trainable biases.

The initial count value ($h_{0}$) has a value of 0 for every sequence and the models are trained in the same manner as the models from Task~1.
The ternary classification accuracy is slightly lower as can be expected with more classes. 
For shorter training sequences, this may be related to the larger number of model parameters relative to the data points.
The accuracy improves with longer training sequences. 
Ternary classification does not show a mean of $a/b$ above $-1$ as can be seen in 
Figure~\ref{fig:histograms_with Bias}.

\section{Conclusions and Future Work}
Although linear RNNs have the theoretical capacity for counting behaviour, previous research has shown that these models often struggle to effectively generalise counting behaviour to long sequences. 
In this study, we present a set of necessary and sufficient conditions that indicate counting behaviour in linear RNNs and provide proof that meeting these conditions is equivalent to counting behaviour. 
To investigate the extent to which these conditions are met, we examine the parameters of models trained on sequences of varying lengths for classification tasks. We use both binary and ternary classification tasks and find that the models do not fully meet these conditions, but do approach them and get closer as the sequence length increases.

There are several potential areas for future work based on these findings. 
One possible research direction is to extend this approach to ReLU RNNs, and LSTMs as far as possible. 
Another option is to devise methods to ensure that the indicator conditions we have identified are met during training in order to improve the generalisation abilities of our models. 


\bibliography{anthology,custom}
\bibliographystyle{acl_natbib}

\appendix

\section{General Counter Machine }\label{apx:definition}

This definition is from \citet{DBLP:journals/corr/abs-2004-06866}.
\begin{dfn}
\normalfont
\label{dfn:general_counter_machine}
(General Counter Machine)
A k-Counter is
a tuple \(\langle \Sigma, Q, q_{0}, u, \delta, F \rangle\) with:
\begin{enumerate}
    \item A finite alphabet \(\Sigma\)
    \item A finite set of states \(Q\)
    \item An initial state \(q_{0}\)
    \item A counter update function
    {\small \[u:\Sigma\times Q\times \{0, 1\}^{k} \rightarrow (\{+m:m\in \mathbb{Z}\}\cup\{\times0\})^{k}\]}
    \item A state transition function
    \[\delta:\Sigma\times Q \times \{0, 1\}^{k} \rightarrow Q\]
    \item An acceptance mask 
    \[F \subseteq Q \times \{0, 1\}\]
\end{enumerate}

\end{dfn}
The counter machine computation is formalised in Definition \ref{dfn:CounterMachineComputation}.
The finite mask of the current state is created using a zero-check function
\(z(\textbf{v})\) for a vector $v$, where:

\begin{equation}
    z(\textbf{v})_{i}=\begin{cases}
			0, & \text{if } v_{i} = 0\\
            1, & \text{otherwise}
		 \end{cases}
		 \label{eq:zero_check}
\end{equation}

\begin{dfn}
\normalfont
(Counter Machine Computation)
\label{dfn:CounterMachineComputation}
\normalfont
Let \(\langle q, \textbf{c}\rangle \in Q \times \mathbb{Z}^{k}\) be a configuration of machine \(M\).
Upon reading input \(x_{t} \in \Sigma\), we define the transition 
\[\langle q, \textbf{c} \rangle \rightarrow_{x_{t}} \langle \delta(x_{t}, q, z(\textbf{c})), u(x_{t}, q, z(\textbf{c}))(\textbf{c})\rangle\]
\end{dfn}

\end{document}